# Knowledge Representation.

*Mikalai Birukou*
*University of Winnipeg, Canada*
*e-mail: mbirukou@io.uwinnipeg.ca*

## Abstract.

This work analyses main features that should be present in knowledge representation. It suggests a model for representation and a way to implement this model in software. Representation takes care of both low-level sensor information and high-level concepts.

## 0. Introduction.

Knowledge representation is, probably, the most important ingredient for developing an AI. A representation is a layer between information accessible from outside world and high-level thinking processes. Without knowledge representation it is impossible to identify what thinking processes are, mainly because representation itself is a substratum for a thought.

The subject of knowledge representation has been "massaged" for a couple of decades already. But we have not been able to find literature where the subject is approached in the same manner as it is done here with the same technical suggestions. Although, we do not claim that all of AI literature has been searched.

In the first section of this work we identify main features that should be present in the knowledge representation, giving a model for representation. The second section suggests how the features described in the first section can be implemented technically. Only basic details are covered. Also we show how the representation can be used. But this last aspect deserves even further investigation.

## 1. A Knowledge Representation Model.

In this section we are going to introduce a model of a knowledge representation. We will concentrate here just on identifying major aspects that should be present in a system that uses knowledge from a real world.

Let us have a look at human brain. A brain is inserted in a body. Functionally it is connected through nerves to receptors of the body. Receptors produce specific signals in a specific form depending on what receptor is. If it is a tactile receptor, it sends signals about temperature and pressure on a particular part of a skin to a nerve, which is responsible for this part of the skin. Central nervous system when receiving signals "knows" that they are coming from tactile sensor on the skin because it sees which nerve carried these signals. Signals due to, for example, high temperature in the nerves coming from different parts of the skin may be of the same shape yet they are not confused because they come from different nerves. Thus, we can say that information comes to brain in two parts, shape of a signal received from a nerve and an origin from were the signal came from.

Shapes of signals that come from receptors change according to conditions on receptors. On another hand information about the origin of a signal is given by what physical nerve is involved into a signal transmission. Therefore, an origin of a signal is determined by "hardware".

Now we make an assumption, which seems to be true. Besides an input from various sensors there are no any other inputs coming into a brain. This suggests that every process in a brain like knowing and understanding should be based on information from sensors.

There may be a lot of examples of how brain "knows" or "understands" some basic concepts that we usually consider trivial common sense knowledge. Here are some questions that one may ask any person and expect an answer.

First situation:
A: "What is one centimetre?"
B: "It's a measure of length or distance."
A: "What is it? I do not understand what you mean."
B: shows two fingers about one centimetre apart – "One centimetre is this much. It is how far my fingers apart. Do you see it? Can you feel it with your fingers?"

This situation often happens when *A* is a kid and *B* is an adult. *A* tries to understand a meaning of the word *centimetre* and reaches understanding when he senses what the word means. Probably, it is a common experience for parents that one has to show all the time what words mean on examples or in games. "This is a dog" – show a real dog or a picture. Otherwise children forget explained meanings. On another hand, childhood is a time when people acquire common sense knowledge. Thus, we may conclude that basic understanding that underlies common sense knowledge is provided by sensor inputs. These sensor inputs provide axiomatic understanding of outside world phenomena.

Second situation:
A: "What is one light year?"
B: "It's a measure of distance. It is how far a photon travels in one year."
A: "How far is it?"
B: "I cannot show. Just extremely far."

In this situation *B* does not have the same understanding of the subject as he does in the first situation. *B* may try to imagine what *light year* is, but he will be doing it on the basis of his understanding of a centimetre. Unfortunately, *B* trying as hard as he can will not be able to really comprehend *"how far is one light year"* because he does not have a direct sensor experience of it. One may say that *B* was told what *light year* is but does not really "understands" it.

Computer program, for example an expert system, can easily produce an answer like *B* did in the second situation. And regular criticism, that programs do not "understand" what they are talking about, will apply to *B* as well. Thus, in the second situation an expert system and a person *B* can be said to be equal in their understanding of a subject that they are talking about. On another hand, in the first situation, where *B* shows understanding, a program will fail because it lacks *B*'s sensor inputs.

Third situation:
A: "Is red the same as green? What is the difference?"
B: points at red and at green – "See the difference."

First of all, this situation will not make sense to colour-blind *A*. So, a person with a little different sense will have a different understanding. Second, this is an example of "how" one "understands" differences between colours. Extrapolating this example, we have that any comparison or other operation with "meanings" on a basic level comes from sensors and sensor inputs.

Up to this point all mentioned examples deal with some concepts that are directly related to our sensors. But does the sensor background help to build complicated pictures (representation) of the outside world. Well, the world is given to us only in sensor experience. Therefore, it is possible to suggest a model, in which any world picture can be build on the basis of a sensor experience.

Let us take any object and think of what we can say about it, what do we know about it, what is it. Take for example a chair. It has particular shape that we may *see* with eyes or *feel* with hands. It has a colour that we may *see*. Some parts of it soft, some are hard. All of these are sensor experience of a given chair. If you are an adult, you may know how different materials look like and how do you feel when you touch them. You were, probably, told that a particular material that *feels* like *this* and *looks* like *this* is called wood. Another material is called leather. So, looking at the chair we also may say what is it maid of. But this information, first, is based on how the chair *looks* like. Second, there is a universal agreement on how to call materials basing on their *look*. Do we know what a material is. No, unless we were *told* that it is made of atoms, because we have no direct experience of what atoms are.

Besides describing different properties of an object we have to mention what happens to an object if you do this or that, does it do anything by itself (animate) when it is left alone, or no (inanimate). This can be called a description of actions or methods that an object may exhibit, or that can be imposed on an object. Say, there are two chairs. They look exactly the same. There is an impression that two chairs are identical. You say jump to one chair, it does not move (normal chair). You say jump to another chair it jumps. Now these two chairs are very different. What makes them different is this second type of description, i.e. action.

We have now two types information for description of any object. Those are static properties and non-static actions. Properties are related directly to how an object is perceived by sensors. Is it possible to say the same about actions? What can be said in general about actions? Let us look at the following examples.

A person looks at chameleon. Chameleon is grey. Person's eyes tell a brain that chameleon is grey. Second later chameleon is green. Person's eyes tell the brain that chameleon is green. If there were no memory about previous state of chameleon's colour in the brain, it would not be able to deduce that colour have changed. But because the brain possesses this memory, it recognizes that the change happened. So, chameleon's activity is represented as a change of chameleon's property (colour).

A person sees a snail. A minute later the snail might be some centimetre away from its previous position. Person's eyes spot new position of the snail. Person's brain recognizes that a new position of the snail is different from the one a minute ago, so, it concludes that a change took place. This particular change is called moving. So, the snail is moving. Again, an action is represented as a change of snail's property (position). If it were a running horse, the whole thinking process would have been immediate because horse's action is more rapid.

These two examples suggest that any action can be represented as a change of object(s)'s properties. This representation is possible only due to a brain being able to remember object(s)'s properties at different times and by placing all encountered object(s)'s states in a correct time sequence(s). Below, we use this representation of actions in our model.

There is yet one more fundamental question to rise. Here we already talked about objects. We inexplicitly assumed that a real world around constitutes of separable objects, elements. For example, we look inside of the room. There are sofas and tables inside. We identify each of those objects, yet they are a part of the whole picture. How brain does this identification? How are we able to distinguish repetitive patterns of a sensory input as a separate object? We will not address these questions here. Instead, we will proceed on how described features of knowledge representation fit into one model and how this model can be implemented on the computer.

Let us now summarize what are the main features of a model of a knowledge representation we have mentioned so far. First, the very basic level of a knowledge representation should be based on sensory information. Second, all categories and manipulations with categories, that one uses, should be based on sensory information. Third, a world is described in terms of objects. Forth, objects are described by their properties and actions that they may exhibit. Fifth, all actions are described as changes of object(s)'s properties. These are five major aspects of our model of a knowledge representation. Is it possible to implement all these aspects in a computer system, and if yes, then how. This question leads us into the second section.

## 2. Implementation of the Model.

To give a design for a model of knowledge representation outlined in the first section, let us stress main features that should be present in the system.

A basis for a knowledge representation in our model is sensory information, handling of sensory inputs. It has been said in the previous section that sensory information consists of two pieces. The first piece is an address of a sensor that generates an input. And the second piece is the signal itself. Therefore, any input from sensor devices should these two things. This can be realized by having a "programming object" (p-object, to avoid confusion with objects from a real life) that one has in usual object oriented languages like C++ or Delphi. So, to be a good package of a sensory input a p-object should have two fields, one with a name or an address of a sensor that produced a signal, and another field with a pointer to a place in a memory where the signal is stored. This p-object can be more complicated, but in this paper we want to mention only basic functionality.

Next element to implement is an object of a real world (it will be referred below just as object). In our model of knowledge representation object is a collection of properties and methods. Can a p-object be an implementation of an object?
P-objects are usually defined by programmer before compiling time. But if system learns some new information about a particular object, then an object should be rebuilt. So, there should be a possibility for a system to change objects during a run time in any way, i.e. add or erase any property or method. Also, not every time the object is loaded into memory all of its components are needed. For example, system creates a picture based on the sentence "A man is walking on a street." The only thing mentioned here about a man is that he is walking. So, the only method for a man, which is necessary to load into memory, is "walking". There is no need to load into limited RAM of the computer a possibility that this man, say, can be married, etc. By not loading all possible information about a man we on one hand we save computer resources, on another hand we are not risking to be overloaded by unnecessary information. If the second sentence is added, "He has an umbrella in his hand", then one has to add to the objects in RAM that a man has a hand, and this hand carries an umbrella. So, only features that are necessary are loaded into computer's memory. Here no new information has been learned about men in general, but we still need dynamically changing in memory objects.

The third main element to implement is an action. We have already argued that any action is a change of one or several objects' properties. So, a p-object that implements an action should contain executable instructions. More so, it is just a pointer to a place in memory with those instructions. Here we may also suggest that partial loading into memory can be helpful. For example have a sentence "Man paints a green wall with red." A system may load all necessary objects and actions here. But should it load all instructions for actions. If the system is not asked to do anything else, or it is told to translate the sentence, and translation here may not need actual instruction for actions, then loading of instructions is a waste. But if the system is asked what will happen with the wall, it will load all of the instructions, because these instructions are real implementations of actions, and will execute them. After execution system will look at new values of wall's properties and will spot the fact that the "colour" now has a different value. So, it might be a good idea also to make it possible to load instructions only when they are needed, i.e. before execution.

Let us also note that actions can be characterized by properties, say "how" action goes, etc. So, we have to leave a possibility for this also.

We suggest having p-object for each object, for each property and for each action. So, talking about programming itself, there should be just three basic classes defined, i.e. a class for objects, a class for properties and a class for actions. An additional class may be defined, which is a dynamical array with appropriate helpful methods. P-object of an object points on array-object(s) that contain properties and actions. P-object of a property has a form of a p-object for handling sensory input. P-object of an action points on a set of instructions that tell what should be done. These instructions should be created and processed automatically, so they should be written in the form of internal script language. The system itself should contain an interpreter for this script language. Script should be mainly design to work with objects, properties and actions, allowing some basic operations. On another hand this script might be useful on higher levels, such as collections of objects or functional description of "thinking" methods, if needed.

The first question to rise now is, using the basic element how is it possible to represent more complicated pictures, not just standalone objects and how the interaction between different objects should be implemented. Implementation should be general, so that it will handle any situations possible.

We have one suggestion for a possible interpretation. We do not prove that it is universal. We just note that we have not found any situation in a real world that will not be describable within a frame we are going to propose. This is a bit pragmatic point of view. So, observation is as follows.

Any action is always initiated by something, subject, and it is always addressed onto other object(s) or on a subject itself.

An implementation of this statement consists of putting information into action about its origin (subject) and its destination (object). If a subject is unknown, then a pointer to subject should be blank. But an action never gives rise to another action without any middle object(s). P-object that implements an object should have an array of outgoing actions and of incoming actions. So, describing any situation we have objects that are in the situations and the actions always between them.

This is it. Any situation can be represented now. Whatever follows below deals with how one can use this representation. How the logic can be implemented. How language understanding can be done. What are operations that one can do with object that will remind us of what we do when we think. And so on.

First thing, that one can do with the given basic elements and the way they are put into bigger pictures, is define complex objects. Let us imagine the following example. We have a full representation of a TV-set, i.e. each detail is carefully described with all active connections between details. Now add a person that presses an on/off button with its finger. Button allows current to flow in some circuits, etc. The picture is pretty complicated. What the person sees is changing picture on the screen. Now, the whole part, which describes TV-set, is connected to the person through actions of pressing a button and showing a picture back. Let now allocate instead of this part of the net just one single object and call it TV-set. Now the incoming and outgoing actions for this object may stay the same or may be identified in a new way also, say pressing a button may be called a turning TV on or off. When there is now need in the internal information about TV-set, this "zooming" or abstraction saves a lot of physical resources.

A lecturer on a public lecture about chess pronounced another example of defining complex objects:

*"…Computer programs that play chess do only searches through all possible combinations of moves. Pretty mechanical. I believe that there is something more to it then looking at movement of a pawn or a castle."* He points on a chess board: *"Look, it is not just a king with a castle standing together with pawns in front. This group is difficult to attack. Try opponent's queen here, … Try a horse there, … Etc."*

So, our lecturer basically starts to try different moves, just as a computer program does, but with one exception. All moves are targeted not on a single piece but on a group that possesses some properties. So, the only difference between a computer program and a man here is that a man makes an abstraction, looking at a group of pieces as if it is one thing. This shows, how it is important to give to a system an ability to make this type of abstraction.

This process can be reversed. From a high abstraction level one may go down to a more detailed level, if necessary.

Let us now suggest at least one way of implementing this abstraction. First have a situation that is already represented in objects and actions that are between objects. Then any part of the net that is connected with the rest of the net by actions will be a complex object. And any part of the net that is connected to the rest of net by objects will be a complex action. These identifications support our main principle that objects are connected directly only to actions and actions directly connected only to objects. There may be other ways that we have not thought of, yet.

Let us now realize a notion of a concept. Say, there is a net that describes Mike, as an object that has two legs, two arms and a head. Now, let there be one more net that represents Jack who has two legs, two arms and a head. The properties, say length of Mike's legs and Jack's legs, can be different. But a net's structure, pattern, is the same. If they are the same, then it is possible to identify a concept and call it "a human". This concept is a particular pattern of our representation net. Adding to this pattern appropriate elements we get two more less general concepts of "a man" and "a woman".

So, the repetitive patterns are concepts. Let us now take in pattern and start to reduce it. The smallest possible nets are single object and single action.

We tell to our system that Peter is a man. Can it tell us if Peter has a head? It knows that Peter is a man. Therefore, system will "shape" a net that represents Peter like the net that represents concept "a man". Then it looks into the net and sees "a head" there. So, it gets logical conclusion that Peter does have a head.

Any representation is, actually, a model of a reality. Therefore, our system should work as a universal modeller. Properties of objects describe static pictures. Actions are executable instructions that describe dynamic pictures.

Is it logical that a green wall will be a different colour after somebody paints it black? How can the system figure it out? There is an object "the wall" that has a property "a colour". We say that somebody paints the wall, but "painting" is an instruction that the object of an action will change its colour to black. Make the system run this instruction. After the run our system checks the value of wall's colour. The wall is black. Bingo! We got a logical answer by modelling the situation.

Let us turn attention to the question can a system with the described representation communicate by means of a natural language.

To every representation of an object or an action we have to attach an additional property, which is an identifier in a natural language. More so, there should be as many identifiers as many languages. Otherwise, if it does not know an appropriate word in French, it should simply use an English one instead. Each object should have its own unique internal for a system identifier, so that objects can be found in a memory, or in a database, etc. But language names are not unique. Therefore, language names should be just labels that are used only *to communicate*, not *to think*. For example, both birds and people do fly. It's just that the ways they do it are different. Therefore, an implementation is different. Although, both human's and birds' actions in a natural language are referred as "flying".

The last aspect in our model we touch in this section is "self" and "self-awareness".

To approach this problem we take for granted the following opinion about humans: "*You are exactly what you believe about yourself.*"

This statement for our system means that "self" is an object that describes the system. But this object should have some additional abilities. It should have some kind of a new type of properties that are *goals* and *anti-goals* (something that system should avoid. Also this object should have an ability to activate processes of thinking and other activity. It should have an ability to compare what it thinks with what is in the reality, so it should keep track of sensory inputs, which is the only connection to the reality. Besides these additional features, "self" is an object like any other object of thinking.

## 3. Concluding remarks.

In first two sections we provided a model of a knowledge representation and a description of basis elements that should be present in the system that realizes described model.

We started our discussion with sensors and sensory inputs. The suggested way for handling sensory input is not restricted to any type of sensors. Therefore, our software system can be potentially used in any robotic system.

It was mentioned in the second section that partial loading of objects saves resources. But if we imagine that human brain operates on the same principle as our model, this partial loading of objects is still crucial for it saves a space in a given volume of a "wet ware" that can be used for allocating more objects providing ability to model more complex environments, which can be an evolutionary progressive step.

In the second section we showed some of manipulations that can be done with objects in the system. These manipulations produced results like normal thinking. There is a lot that can be done with objects, and this should be a matter for the following research.